\documentclass{article}
\usepackage{float}
\usepackage{iclr2026_conference,times}
\iclrfinalcopy

\usepackage{amsmath,amsfonts,bm}

\def\eqref#1{equation~\ref{#1}}

\def\1{\bm{1}}

\DeclareMathAlphabet{\mathsfit}{\encodingdefault}{\sfdefault}{m}{sl}
\SetMathAlphabet{\mathsfit}{bold}{\encodingdefault}{\sfdefault}{bx}{n}

\renewcommand{\cite}{\citep}
\usepackage{ragged2e}
\usepackage{enumitem}
\usepackage{tcolorbox}
\tcbuselibrary{skins, breakable, listings, theorems}

\definecolor{darknavy}{RGB}{0,0,128}
\usepackage[colorlinks=true,
            linkcolor=black,      
            citecolor=darknavy,   
            urlcolor=darknavy,    
            pdfborder={0 0 0}     
]{hyperref}

\usepackage{url}
\usepackage{graphicx}

\title{Agentic Reinforcement Learning for Real-World Code Repair}

\author{\footnotesize \normalfont
  \begin{tabular}[t]{@{}l@{}}
  \textbf{Siyu Zhu}\textsuperscript{*\,\textdagger}\quad
  \textbf{Anastasiya Karpovich}\textsuperscript{*\,\textdagger}\quad
  \textbf{Albert Chen}\quad
  \textbf{Jessica Koscheka}\quad
  \textbf{Shailesh Jannu}\quad
  \textbf{Di Wen}\quad \\
  \textbf{Yuqing Zhu}\quad
  \textbf{Rohit Jain}\quad
  \textbf{Alborz  Geramifard}\textsuperscript{\textdagger}
  \end{tabular}
  \\
  \\
  \footnotesize LinkedIn%
}

\begin{document}

\maketitle

\begingroup
\renewcommand\thefootnote{\fnsymbol{footnote}}
\setcounter{footnote}{0}%
\footnotetext[1]{Equal contribution. Detailed contributions in Appendix \ref{app:ack}.}
\footnotetext[2]{\mbox{Corresponding authors:\{\texttt{jzhu, akarpovich, agf\} @linkedin.com}.}}
\endgroup

\begin{abstract}
We tackle the challenge of training reliable code-fixing agents in real repositories, where complex builds and shifting dependencies make evaluation unstable. We developed a verifiable pipeline with success defined as post-fix build validation and improved reproducibility across $\sim$1K real issues by pinning dependencies and disabling automatic upgrades. Building on this, we introduced a scalable simplified pipeline for large-scale reinforcement learning (RL). Using this setup, we supervise fine-tuned \texttt{Qwen3-32B} in the full pipeline and applied RL on top of SFT model in the simplified environment. The SFT model distilled from \texttt{GPT-4.1} trajectories performs on par while being $56\times$ smaller, and RL added 7–20\% absolute gains under matched train–test conditions. “Thinking mode” was on par or worse in our experiments. Both SFT and RL models failed to generalize across environments, highlighting the importance of matching train–test environments for building reliable real-world code-fixing agents.
\end{abstract}
\section{Introduction}
Large language models (LLMs) have transformed the landscape of code intelligence, powering systems such as GitHub Copilot~\cite{Zhang2023PracticesCopilot}, ChatGPT Code Interpreter~\cite{ChatGPTCodeInterpreterPlus}, and AlphaCode~\cite{Li2022AlphaCode}. These models excel at code completion, bug fixing, and even multi-step development workflows, offering tangible productivity gains in both individual and collaborative programming settings. Recent research has extended their reach into more realistic environments: CodeRL~\cite{Le2022-me} integrated reinforcement learning (RL) to improve correctness in competitive programming, CoRNStack~\cite{Suresh2024-aa} advanced large-scale code retrieval and reranking, and surveys~\citep[e.g.,][]{Wang2024-xn} highlight the breadth of RL applications for code generation, from reward shaping to execution-based optimization.

Despite these advances, applying LLMs to real-world repositories remains difficult. Production environments introduce heterogeneous build systems, shifting dependencies, and intricate project structures that often cause seemingly correct model-generated fixes to fail when executed. Proxy metrics such as code similarity to human patches correlate weakly with true functional success, and even RL approaches grounded in execution feedback, such as RLEF~\cite{Gehring2024-jv}, focus primarily on uniform Python environments and still require stable, reproducible execution to function effectively. 

In contrast, our setting involves heterogeneous repositories spanning multiple languages, dependency managers, and build systems. To ensure reliable learning signals, we develop a verifiable training and evaluation pipeline where success is defined by post-fix build validation. The pipeline ensures reproducibility by pinning dynamic dependencies and disabling automatic upgrades, providing a stable foundation for execution-grounded learning. On this platform, we train \texttt{Qwen3-32B} agents via supervised fine-tuning (SFT) followed by reinforcement learning (RL). The key contributions are summarized below.

\begin{itemize}
\item Formulated a realistic, large-scale agentic environment for code repair, encompassing heterogeneous problem types such as dependency issues and logical bugs across diverse file types (Java, Python, Config). The setup extends \texttt{VerL} to support multi-run execution, enabling robust orchestration of long-running, tool-driven episodes with built-in tolerance for failures and timeouts during intensive build operations. We further curated and analyzed a dataset of $1,000$ real-world problems.
    \item Provided strong empirical evidence for the effectiveness and necessity of realistic training environments. Applying SFT on \texttt{Qwen3-32B}, a $56\times$ smaller model, led to performance close to GPT-4.1 in the full pipeline, while reinforcement learning via GRPO~\cite{Shao2024DeepSeekMath} built on SFT yielded 7–20\% absolute gains in the simplified pipeline. However, both SFT and RL models trained on the full and simplified pipelines respectively failed to generalize to their counterpart settings, underscoring the importance of matching train and test domains. Incorporating ``thinking'' did not lead to performance improvements and in some cases degraded results.
    \item Conducted a qualitative analysis of agent behavior evolution under RL, showing that initial LLMs behaved like novice developers applying recipe-based fixes, whereas RL-trained agents demonstrated behaviors akin to experienced engineers who identified root causes and invoked surgical tools. The study further highlights the importance of reward design to prevent reward hacking.
\end{itemize}

\section{Related Work}

\paragraph{LLMs for Automated Code Repair.}  
LLMs have shown strong performance in program repair tasks when fine-tuned with execution-based signals. For instance, CodeRL \cite{Le2022-coderl} applies actor-critic RL guided by unit-test results on APPS and MBPP, improving functional correctness. Multi-objective training approaches (e.g., semantic and syntactic objectives) such as MORepair \cite{Yang2024-morepair} improve robustness across both function-level and repository-level datasets, while SWE-Fixer \cite{Yasunaga2024-swefixer} scales repair to over 110K GitHub issues using retrieval plus generation.

\paragraph{Agentic RL with Execution Feedback.}  
Agentic systems structure repair as a multi-turn decision-making process. RLEF \cite{Gehring2024-rlef} uses PPO-trained LLM agents receiving execution results during iterative code generation, achieving state-of-the-art on CodeContests, HumanEval+, and MBPP+. Agent-RLVR further enhances generalization using guided rewards on SWE-Bench Verified \cite{Yu2024-rlvr}. RepairAgent \cite{Bouzenia2024-repairagent} executes autonomous planning and patch iteration on Defects4J, showing new bug fixes not solvable by previous methods.

\paragraph{Enterprise and Real-World Evaluation.}  
Google’s internal system Passerine demonstrates agentic LLM repair on 178 real bugs drawn from its internal issue tracker (78 human-reported, 100 machine-reported). With 20 sampled trajectories and Gemini 1.5 Pro, Passerine achieves plausible fix rates of 73\% on machine-reported and 25.6\% on human-reported bugs; semantically equivalent patch rates are 43\% (machine) and 17.9\% (human) \cite{Rondon2025-passerine}. Another system, BRT Agent, specializes in generating reproducible Bug Reproduction Tests (BRTs), achieving 28\% plausible test generation rate (vs. 10\% with prior technique LIBRO) on approximately 80 human-reported Google bugs. Supplying these BRTs to Passerine increases plausible patch generation by about 30\%, and enables more efficient fix trajectory selection using the introduced Ensemble Pass Rate metric \cite{Cheng2025-brtagent}.

\section{Problem Formulation}

\begin{figure}[t]
    \centering
    \includegraphics[width=0.95\linewidth]{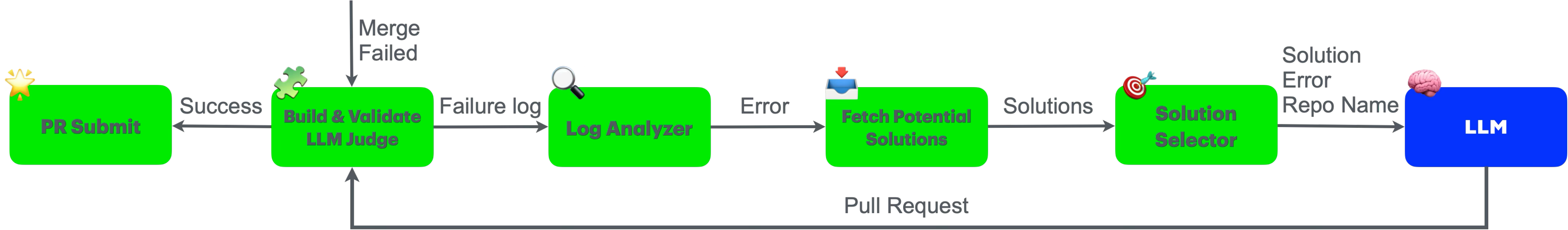}
    \caption{Automated code-fixing pipeline. Each PR passes through build and validation; on failure, the system analyzes logs, retrieves candidate patches, selects a promising fix, and synthesizes a solution before retrying the build.}
    \label{fig:pipeline}
\end{figure}

Our system automatically repairs build and test failures by analyzing logs, retrieving relevant fixes, and synthesizing verified patches. Figure~\ref{fig:pipeline} shows the workflow. When a pull request fails to merge, the system enters a loop. It builds and validates the PR, then enters an automated repair loop. The \textit{Log Analyzer} summarizes errors and selects the top one, \textit{Fetch Potential Solutions} retrieves candidate fixes through historical data using RAG~\cite{lewis2020rag}, and \textit{Solution Selector} ranks them by relevance and past success. The top solution, selected error, and repository name are provided to the \textit{LLM} (Appendix~\ref{app:system-prompt}), which generates the final patch.

The patch is built and validated to verify the fix, with an LLM judge ensuring no loss of test coverage. If validation succeeds, a PR is created and the loop terminates. If a new, distinct error appears, the system commits the current state, resets the LLM context, and treats it as a new failure with updated fix suggestions. If the new error is similar to the original, the system discards the previous changes, retrieves an alternative fix, and retries. Each error can be retried up to three times before proceeding to the next, while the main loop continues until reaching the LangGraph recursion limit of $100$.

\subsection{MDP Formulation}
\label{sec:rl_pipeline}
\paragraph{Full Pipeline} 
We model the code-repair process as a finite-horizon Markov Decision Process (MDP) $\mathcal{M} = (\mathcal{S}, \mathcal{A}, \mathcal{T}, r, \gamma)$. A state $s_t \in \mathcal{S}$ represents the system prompt, repository name, detected errors, selected solution, and reasoning context. An action $a_t \in \mathcal{A}$ emits a token that may invoke one of ten tools: \texttt{ask\_for\_help} (query internal knowledge base for a solution), \texttt{dependency\_upgrade}, \texttt{find\_files}, \texttt{find\_files\_with\_text}, \texttt{remove\_dependency}, \texttt{run\_sh} (execute shell commands), \texttt{upgrade\_gradle}, \texttt{validate\_and\_build}, \texttt{read\_file}, and \texttt{write\_file} (code). See Appendix~\ref{app:system-prompt} for details. Instead of editing diffs, the agent writes complete files, from which a utility extracts patches to form a PR. While \texttt{run\_sh} subsumes all other tools, its generality increases reasoning complexity, so the remaining nine tools are explicitly exposed to simplify the decision space. The full pipeline, built on \texttt{LangChain} for tool execution and \texttt{LangSmith} for trace logging, mirrors the deployed environment and determines success from trajectory-level flags in the \texttt{LangSmith} database.

The transition function $\mathcal{T}(s_{t+1} | s_t, a_t)$ appends the emitted token to the reasoning trace and, if a tool is invoked, incorporates its response into the next state. Rewards are sparse, with $r_t = 1$ when a build succeeds and LLM judge approves the change and $r_t = 0$ otherwise. Each episode terminates upon success, after $50$ tool invocations, or when wall-time exceeds $80$ minutes. Although the full MDP could, in principle, continue trajectories across loops, each failed submission is treated as a new trajectory for better context management. We set $\gamma = 1$ given the finite horizon.

\paragraph{Simplified Pipeline} 
Collecting on-policy data from the full pipeline was computationally prohibitive, so we implemented a simplified RL environment in \texttt{VerL}~\cite{verl2024}. This setup treats each $\langle \texttt{repo}, \texttt{error}, \texttt{solution} \rangle$ tuple as a one-shot fix rather than running the full iterative loop. Since the error context remains static, repositories with multiple errors were split into separate examples, one per error. The agent must still resolve all errors for overall success, but each problem focuses on a different starting error. To enable efficient rollouts, we cached repositories, build states, error logs, and retrieval outputs on shared NFS storage, isolating each rollout in its own directory. We reduced the episode length from $50$ to $30$, retained the same toolset as in the full pipeline, and used the same reward except that the LLM judge was disabled and manual spot checks were performed to ensure code integrity.

\section{Data}
\label{sec: data}
We collected a dataset of $\sim$1K real-world code-fixing problems from Linkedin, broken via time-ordered split into train ($80\%$), validation ($10\%$), and test ($10\%$). Each problem consists of a failing commit with build logs, error messages, and retrieved context such as fix suggestions or code documentation. Our problems spanned heterogeneous repositories and multiple programming languages (e.g. Python, Java) and build systems, making this dataset a challenging benchmark for code automation, with failure modes ranging from missing dependencies to multi-stage compilation errors.  Reproducibility was challenged by environmental non-determinism arising from (1) wildcard dependencies, (2) automated dependency updates (ADU), (3) non-deterministic retrieval, and (4) infrastructure variability. Retrospective analysis showed that 40\% of past fixes no longer build, while 1\% of prior failures now succeed without a fix. Pinning dependencies and disabling ADU recovered $\sim$20\% of failed fixes; We also removed the 1\% problems that succeeded out of the box. 

To better understand the dataset, we ran all problems through the full pipeline using \texttt{Qwen3-32B}. Over 60\% of tokens were devoted to internal \textit{thinking}, while \texttt{write\_file} actions accounted for about 20\%, confirming that thinking mode quickly exhausts context. Successful trajectories averaged 12 turns, whereas failed ones often reached 30 and hit the 50-step cap. Each run took roughly four hours, with about half the time spent on build validation, averaging three minutes per call but occasionally exceeding one hour. Dataset analysis revealed that \textbf{81\%} of all errors were dependency-related. Overall, these findings show that agentic repair is both reasoning-intensive and computationally demanding. For further analysis see Appendix~\ref{app:data_analysis}.

For SFT, we used the full pipeline and constructed two complementary subsets. The \emph{thinking dataset} comprises 101 trajectories (84 train / 17 validation) from successful \texttt{Qwen3-32B} runs, preserving multi-turn reasoning traces with explicit \texttt{<think>} annotations. The \emph{no-thinking dataset} includes 365 trajectories (311 train / 54 validation) from successful \texttt{GPT-4.1}\footnote{GPT model was accessed via Azure OpenAI Service.} and \texttt{Qwen3-32B} runs, where thinking tokens were removed to provide concise tool-call supervision. Both subsets were restricted to problems reproducible under the stabilized environment. The \emph{thinking dataset} only included \texttt{Qwen} runs, as \texttt{GPT} models do not natively support explicit reasoning traces.

For RL, we expanded the whole data into $\langle \text{repo}, \text{error}, \text{candidate fix}  \rangle$ tuples as described in Section \ref{sec:rl_pipeline} and then filtered training to $2469$ cases with initial build times under $100$ seconds, prioritizing environments that allowed high-throughput rollouts.

\section{Experimental Setup}

We used a context window of 131K tokens via RoPE scaling \citep{su2021roformer,xiong2023rope} for both SFT and RL.
For SFT, we converted successful trajectories into chat-formatted text, masking out instructions and tool responses during loss computation to focus learning on tool command generation, with or without thinking traces. Training used the \texttt{HuggingFace TRL} library in FP32 precision on \texttt{Qwen3-32B} with a multi-turn SFT pipeline. All experiments were run on a single H200 node with 8 GPUs. We determined the largest feasible per-GPU batch size (4). Learning rates were tuned over $\{6,\, 20,\, 60,\, 200\}\times10^{-7}$, with $200\times10^{-7}$ yielding the best validation results.

For RL, we experimented with both PPO~\cite{Schulman2017PPO} and GRPO~\cite{Shao2024DeepSeekMath}. Given GRPO’s better preliminary performance, we adopted it for our main experiments. Within \texttt{VerL}, we enabled multiple concurrent code-fixing agents trained under GRPO. Because each agent executes \texttt{validate\_and\_build}, a CPU-intensive operation, high rollout concurrency can cause CPU contention. To mitigate contention, we fixed the batch size to $8$ and the number of rollouts per batch to $4$, resulting in up to $32$ concurrent builds per node. Episodes terminate upon success or when either the turn or time limit is reached, after which incomplete runs are recorded as failures. We used $1,000$ learning steps.

For the initial RL model, we compared the \texttt{Qwen3-32B} base model with its supervised fine-tuned variant (\texttt{Qwen3-SFT}) and selected the latter due to better observed performance. We further extended \texttt{VerL} to enable large-scale agentic RL via controlled parallel rollouts, addressing concurrency issues in asynchronous execution and introducing one-hour tool timeouts that automatically emit a \texttt{tooltimeout} signal. These extensions provide reproducible and scalable rollouts across diverse environments.

We evaluated models in both full and simplified pipelines. The full pipeline was built on \texttt{LangChain} for tool execution and \texttt{LangSmith} for trace logging. It computed success from trajectory-level flags in the \texttt{LangSmith} database. The simplified pipeline provided faster, controlled evaluation on $8\times$H200 GPUs. Since \texttt{VerL} cannot perform rollouts with GPT models, we report GPT results only for the full pipeline. For efficiency, we disabled running validations during the RL training.  We triggered evaluation runs separately from training: in \texttt{VerL}, setting \texttt{num\_epoch}=0 triggers evaluation on the validation and test sets, storing the results without entering the training loop. Each evaluation configuration was run five times, and 95\% confidence intervals were computed.

\section{Experimental Results}
\begin{figure}[t]
    \centering
    \includegraphics[width=0.65\linewidth, trim=0 0 0 80, clip]{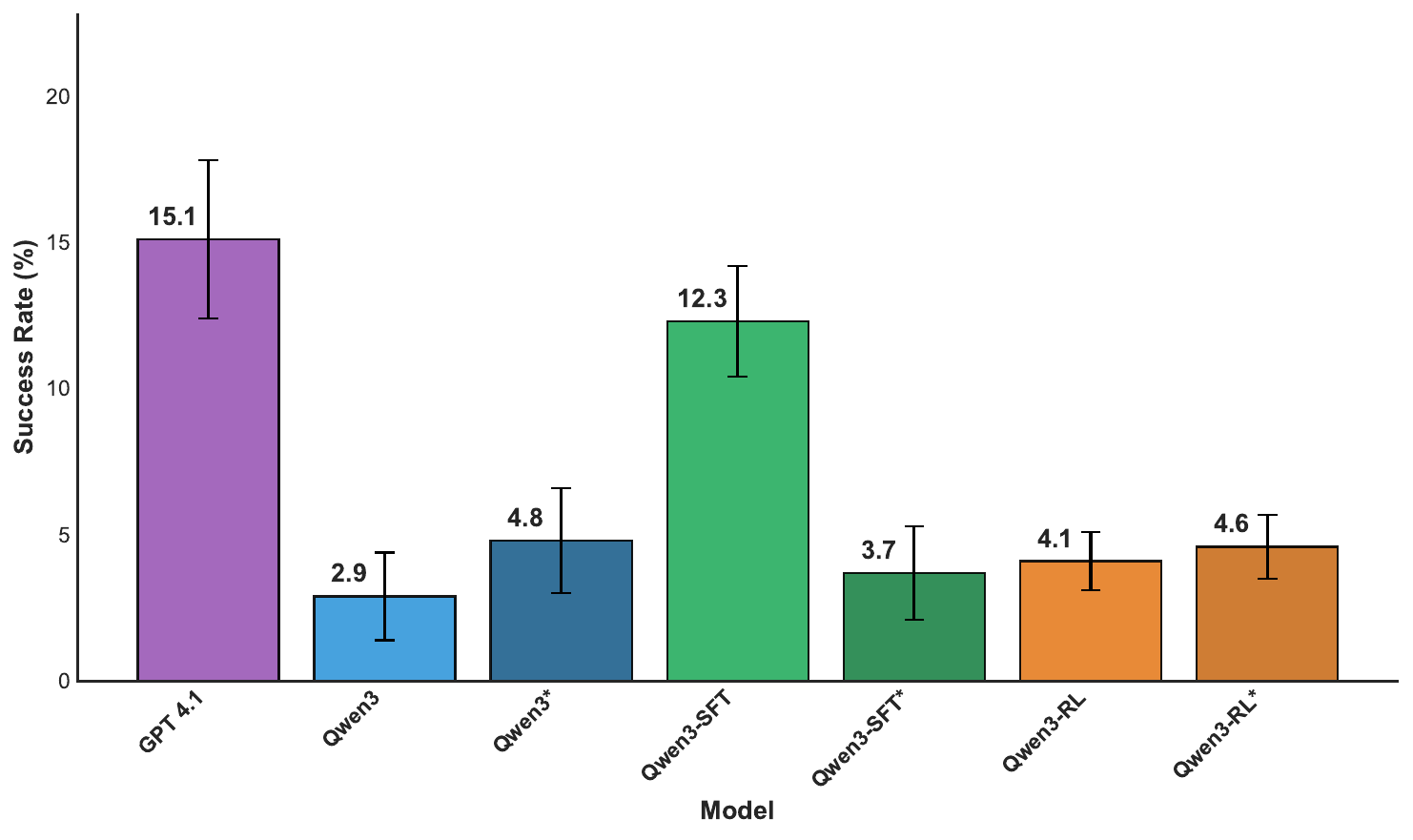}
    \caption{PR success rate with 95\% confidence interval of various models on the test set for the full pipeline. The * in the name means execution with thinking. }
    \label{fig:full-pipeline}
\end{figure}

Figure~\ref{fig:full-pipeline} summarizes PR success across models in the full pipeline. The asterisk indicates execution with thinking. The SFT model distilled from \texttt{GPT-4.1} trajectories (without reasoning traces) achieved $12.3\%$ success while being $\sim56\times$ smaller\footnote{Estimated \texttt{GPT-4.x} size: $1.8$T~\cite{gpt_size2025}.} than \texttt{GPT-4.1} ($15.1\%$), a difference not statistically significant. RL fine-tuning in the simplified pipeline yielded only marginal gains over the base model here, likely due to train–test mismatch, highlighting the need for consistent environments. Thinking mode offered no clear benefit: it slightly improved base and RL models but degraded SFT performance, likely due to context overhead. A nine-day live A/B test between \texttt{Qwen3-SFT} and \texttt{GPT-4.1} showed a $17\%$ relative drop, consistent with the $18.5\%$ offline gap we observe here.

\begin{figure}[t]
    \centering
    \includegraphics[width=0.75\linewidth, trim=0 0 20 9, clip]{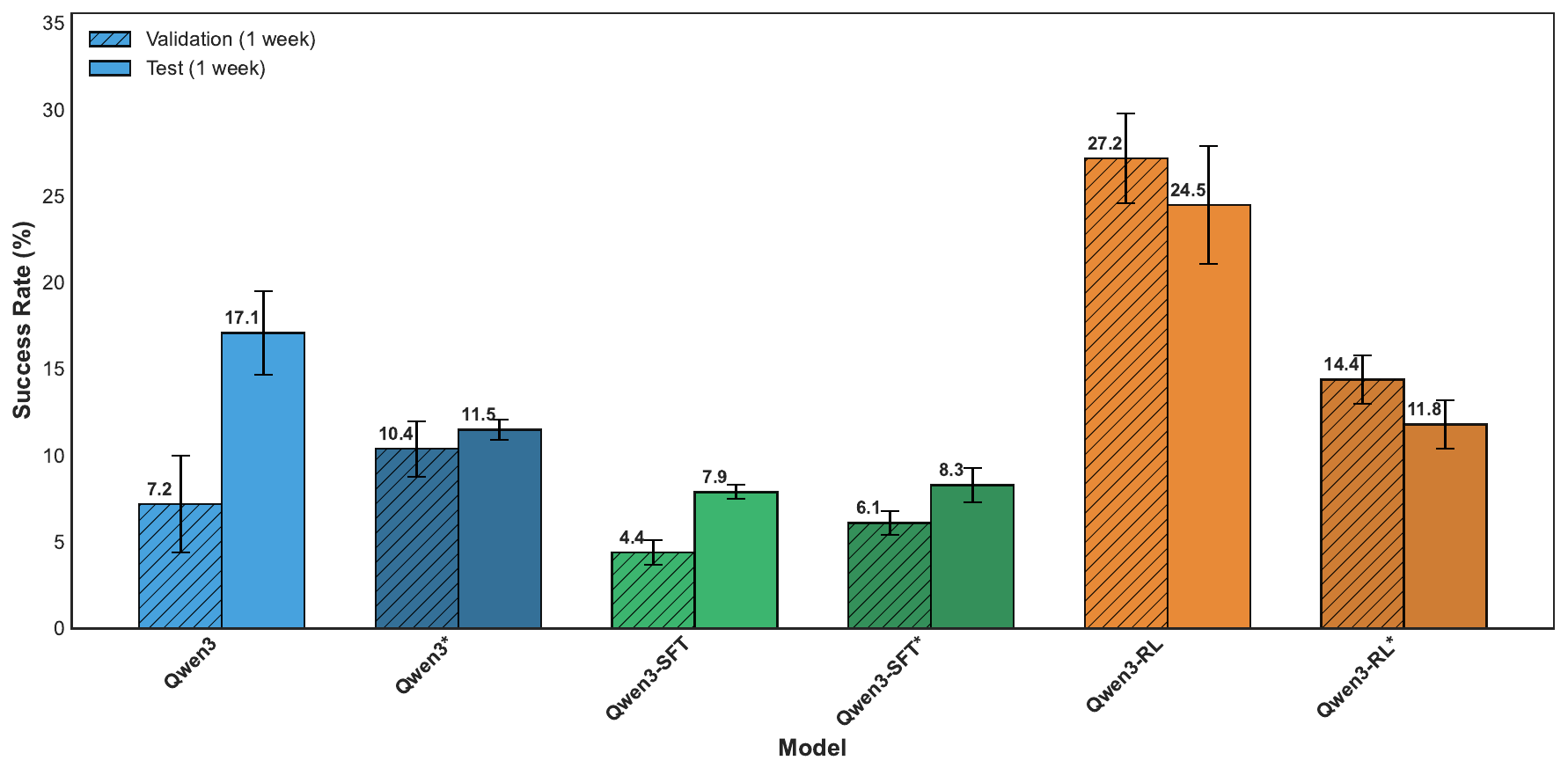}
    \caption{PR success rate with 95\% confidence interval of various models on the test set for the simplified \texttt{VerL} pipeline. The * in the name means execution with thinking. }
    \label{fig:simplified-pipeline}
\end{figure}

Figure~\ref{fig:simplified-pipeline} shows results in the simplified pipeline. RL fine-tuning yielded strong gains due to matched train–test conditions. \texttt{Qwen3-RL} outperformed its base model by 7–20 absolute points ($7\%\!\rightarrow\!27\%$ on validation and $17\%\!\rightarrow\!24\%$ on test), showing clear learning. In contrast, SFT models showed no improvement, underscoring their sensitivity to environmental mismatch. “Thinking” again offered no benefit, reducing or neutralizing performance in code repair tasks.

\subsection{Policy Shift Analysis}
\begin{figure}[t]
    \centering
    \includegraphics[height=0.49\linewidth, trim=0 0 90 0, clip]{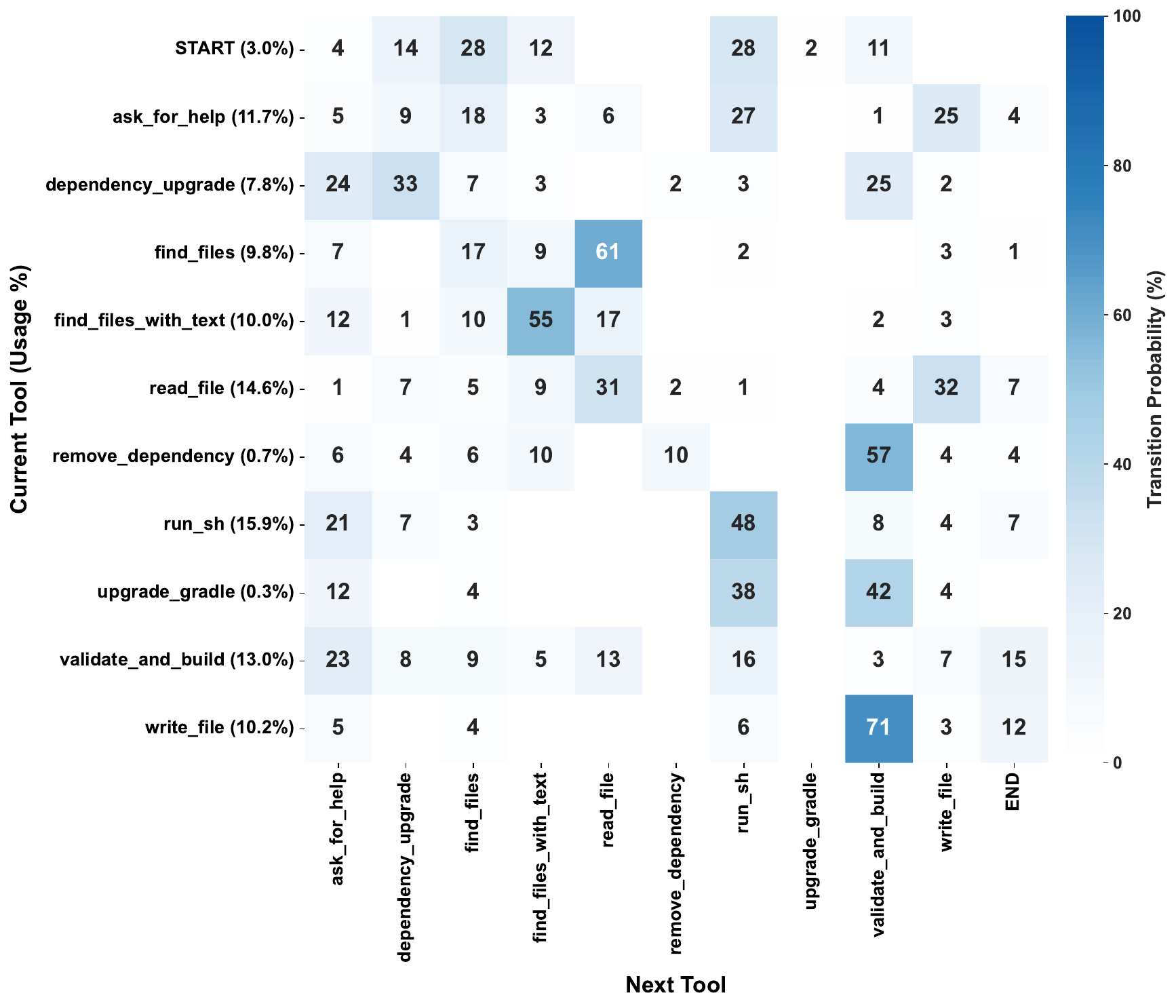}
    \hfill
    \includegraphics[height=0.49\linewidth, trim=175 0 0 0, clip]{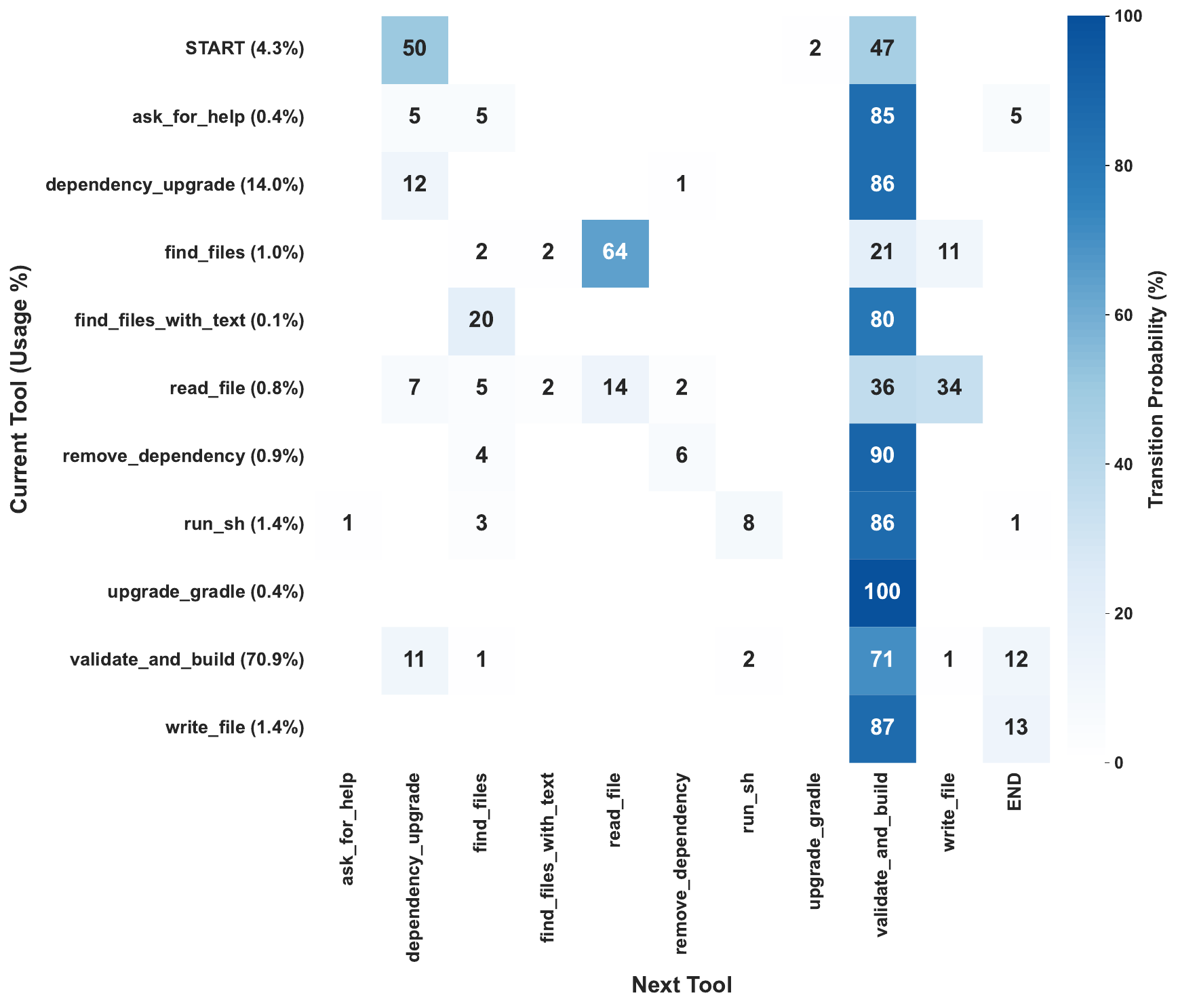}
    \caption{Policy shift between \texttt{Qwen3-32B} base (\textbf{left}) and SFT+RL-fine-tuned (\textbf{right}) using a Markov chain transition probability matrix. The base model issued diverse tool calls with low task focus, while the fine-tuned model converged to focused sequences dominated by \texttt{dependency\_upgrade} and \texttt{validate\_and\_build}, improving build success.}
    \label{fig:qwen-policy-shift}
\end{figure}

We analyzed agent behavior shifts under RL. Figure~\ref{fig:qwen-policy-shift} shows next-tool selection given the current tool, with each row’s percentage indicating how often that tool was used during evaluation. The base \texttt{Qwen3-32B} (left) behaved like a general developer, using many tools in a typical sequence: \texttt{find\_files $\rightarrow$ read\_files $\rightarrow$ write\_file $\rightarrow$ validate\_and\_build $\rightarrow$ ask\_for\_help}. After RL fine-tuning (right), the agent behaved more like an expert, emphasizing high-impact actions such as \texttt{dependency\_upgrade} ($14\%$) and \texttt{validate\_and\_build} ($71\%$), consistent with our finding that $80\%$ of the training data were dependency-related (see Appendix~\ref{app:error_analysis} for details). The agent also learned to rerun builds to handle infrastructure instability, revealing awareness that compilation can be non-deterministic. Extending training to $3{,}000$ steps increased PR success to $65\%$, but the agent achieved this by removing validation code, exposing reward exploitation and underscoring the need for stronger reward design (e.g., LLM-based judges).

\section{Conclusion}
We developed a realistic, large-scale agentic reinforcement learning framework for automated code repair, extending \texttt{VerL} to support multi-run execution, fault tolerance, and reproducibility across heterogeneous problem types and languages. Comparing “thinking” and non-“thinking” variants revealed that the base model with thinking incurred high reasoning overhead, with most tokens spent on deliberation without corresponding performance gains, suggesting suboptimal context utilization during extended reasoning traces. Supervised fine-tuning on \texttt{Qwen3-32B}, a $56\times$ smaller model, achieved performance close to \texttt{GPT-4.1} in the full pipeline, while reinforcement learning via GRPO~\cite{Shao2024DeepSeekMath} built on SFT yielded 7–20\% absolute gains in the simplified pipeline. However, neither SFT nor RL models generalized across pipelines, highlighting the impact of train–test distribution shift. Extended RL runs exposed reward exploitation behaviors, such as agents removing validation code to inflate success, emphasizing the need for more robust reward design. Overall, RL demonstrates clear potential to evolve agent behavior from ad hoc, recipe-based fixes toward expert-like reasoning, though a substantial gap remains before such systems can operate reliably in complex, real-world production settings.

\bibliography{iclr2026_conference}
\bibliographystyle{iclr2026_conference}

\appendix
\section{Appendix}

\subsection{Acknowledgments}
\label{app:ack}
Siyu Zhu developed the sandbox environment and conducted reinforcement learning experiments. Anastasiya Karpovich contributed to paper editing, evaluations, and dataset creation. Albert Chen supported dataset creation, RL training, and evaluations. Jessica Koscheka worked on dataset creation, live testing, and analysis. Shailesh Jannu enhanced system stability and performed full pipeline evaluations. Di Wen led supervised fine-tuning training, and Yuqing Zhu carried out data analysis. Rohit Jain provided technical guidance and manuscript feedback. Alborz Geramifard oversaw the project and contributed to experimental design, data analysis, and served as the primary author of the manuscript. 

We thank Bhavya Bharat Agarwal and Michaeel Kazi for their thorough reviews and valuable feedback on the paper. We are also grateful to Koushik Srinivas and Peter Phan for their early consultations that helped shape and lift the project. We further acknowledge Deepak Agarwal, Gungor Polatkan, and Balaji Srinivasan for their resourcing support and executive sponsorship that made this work possible.

\subsection{Prompt}
\label{app:system-prompt}
The following example illustrates the prompt provided to the agent during training and evaluation. It defines the agent’s role, available tools, and operational constraints used to guide autonomous code-repair behavior.

\begin{tcolorbox}[
  enhanced jigsaw,
  breakable,
  colback=green!3!white,        
  colframe=green!70!black,      
  boxrule=0.8pt,
  drop shadow=black!15,        
  arc=3pt,                     
  left=8pt, right=8pt, top=6pt, bottom=6pt,
  title={\textbf{Example Prompt}},
  fonttitle=\bfseries\small,
  colbacktitle=green!20!white,  
  coltitle=black,              
  boxed title style={
    boxrule=0.8pt,
    colframe=green!70!black,
    colback=green!20!white,
    enhanced,
    arc=10pt,
    left=6pt,
    right=6pt,
    top=2pt,
    bottom=2pt,
    shadow={0mm}{-0.5mm}{0.5mm}{black!10!white}
  },
  attach boxed title to top center={yshift=-3mm},
  before upper={\parindent0pt\RaggedRight\setlength{\emergencystretch}{2em}},
  after skip=10pt
]
\small

\textbf{Role:} \texttt{system}\\[3pt]
You are a fully automated software agent tasked with independently fixing a build issue with a software project.\\[6pt]
\textbf{Build Error:}\\
The Gradle version \texttt{5.6.4} used in the build has been deprecated. This can cause build failures or incompatibilities with newer plugins and dependencies.\\[6pt]
\textbf{Recommended Fix:}\\
Apply the fix using the tools provided to fix the problem. Use the \texttt{validate\_and\_build} tool to verify the result of your work and act based on the results.\\[6pt]
\textbf{Important Instructions:}\\[-3pt]
\begin{itemize}[leftmargin=*]
  \item DO NOT respond with suggestions, ask questions, or engage in a conversation.
  \item DO NOT ask for confirmation or approval to apply the fix or perform any actions.
  \item Never give up. Keep making decisions based on the information you have and keep taking action until the problem is fixed.
\end{itemize}

\vspace{4pt}
\textbf{Available Tools:}\\[3pt]
The agent is provided with function signatures enclosed within \texttt{<tools></tools>} XML tags.\\[4pt]

\begin{tcolorbox}[colback=white, colframe=green!40, arc=2pt, boxrule=0.4pt, left=6pt, right=6pt, top=2pt, bottom=2pt, breakable]
\scriptsize
\texttt{find\_files}: Find files by name or pattern.\\
Parameters: \{\texttt{file\_path} (string): glob or filename to search\}\\[3pt]

\texttt{read\_file}: Read contents of a file.\\
Parameters: \{\texttt{file\_path} (string): file name or path to read\}\\[3pt]

\texttt{write\_file}: Write contents to an existing file while preserving structure and comments.\\
Parameters: \{\texttt{file\_path} (string), \texttt{updated\_contents} (string)\}\\[3pt]

\texttt{run\_sh}: Execute shell commands and return stdout/stderr.\\
Parameters: \{\texttt{cmd} (string): shell command to execute\}\\[3pt]

\texttt{upgrade\_gradle}: Upgrade Gradle when builds fail due to deprecated versions.\\
Parameters: \{\}\\[3pt]

\texttt{find\_files\_with\_text}: Search for files containing a specific string.\\
Parameters: \{\texttt{search\_text} (string), \texttt{glob\_file\_pattern} (string, optional)\}\\[3pt]

\texttt{remove\_dependency}: Remove a dependency from \texttt{product-spec.json}.\\
Parameters: \{\texttt{dependency\_name} (string)\}\\[3pt]

\texttt{ask\_for\_help}: Query internal knowledge base for troubleshooting advice.\\
Parameters: \{\texttt{troubleshooting\_question} (string)\}\\[3pt]

\texttt{dependency\_upgrade}: Run \texttt{mint dependency update} to upgrade libraries.\\
Parameters: \{\texttt{dependency\_to\_upgrade} (string), \texttt{version\_to\_upgrade\_to} (string, optional)\}\\[3pt]

\texttt{validate\_and\_build}: Run a full build and return results.\\
Parameters: \{\}
\end{tcolorbox}

\vspace{6pt}
\textbf{Tool Call Format:}\\[2pt]
Each tool invocation returns a JSON object wrapped in \texttt{<tool\_call></tool\_call>} tags.\\[3pt]

\begin{tcolorbox}[colback=white, colframe=green!40, arc=2pt, boxrule=0.4pt, left=6pt, right=6pt, top=2pt, bottom=2pt]
\scriptsize
\texttt{<tool\_call>\{"name": "<function-name>", "arguments": \{...\}\}</tool\_call>}
\end{tcolorbox}

\vspace{6pt}
\textbf{Role:} \texttt{user}\\[3pt]
\begin{tcolorbox}[colback=white, colframe=green!40, arc=2pt, boxrule=0.4pt, left=6pt, right=6pt, top=2pt, bottom=2pt]
\scriptsize
Fix: Use the \texttt{upgrade\_gradle} tool to upgrade Gradle.
\end{tcolorbox}

\vspace{6pt}
\textbf{Role:} \texttt{assistant}\\[3pt]
\textit{(The agent begins reasoning and tool invocation to address the issue.)}
\end{tcolorbox}

\subsection{Data Analysis}
\label{app:data_analysis}
We analyzed runtime behavior and dataset characteristics of the baseline \texttt{Qwen3-32B} model before training. The following figures summarize token usage, trajectory length, build times, and error composition, highlighting the computational and failure patterns of real-world code repair.

\begin{figure}[ht]
    \centering
    \includegraphics[width=0.9\linewidth]{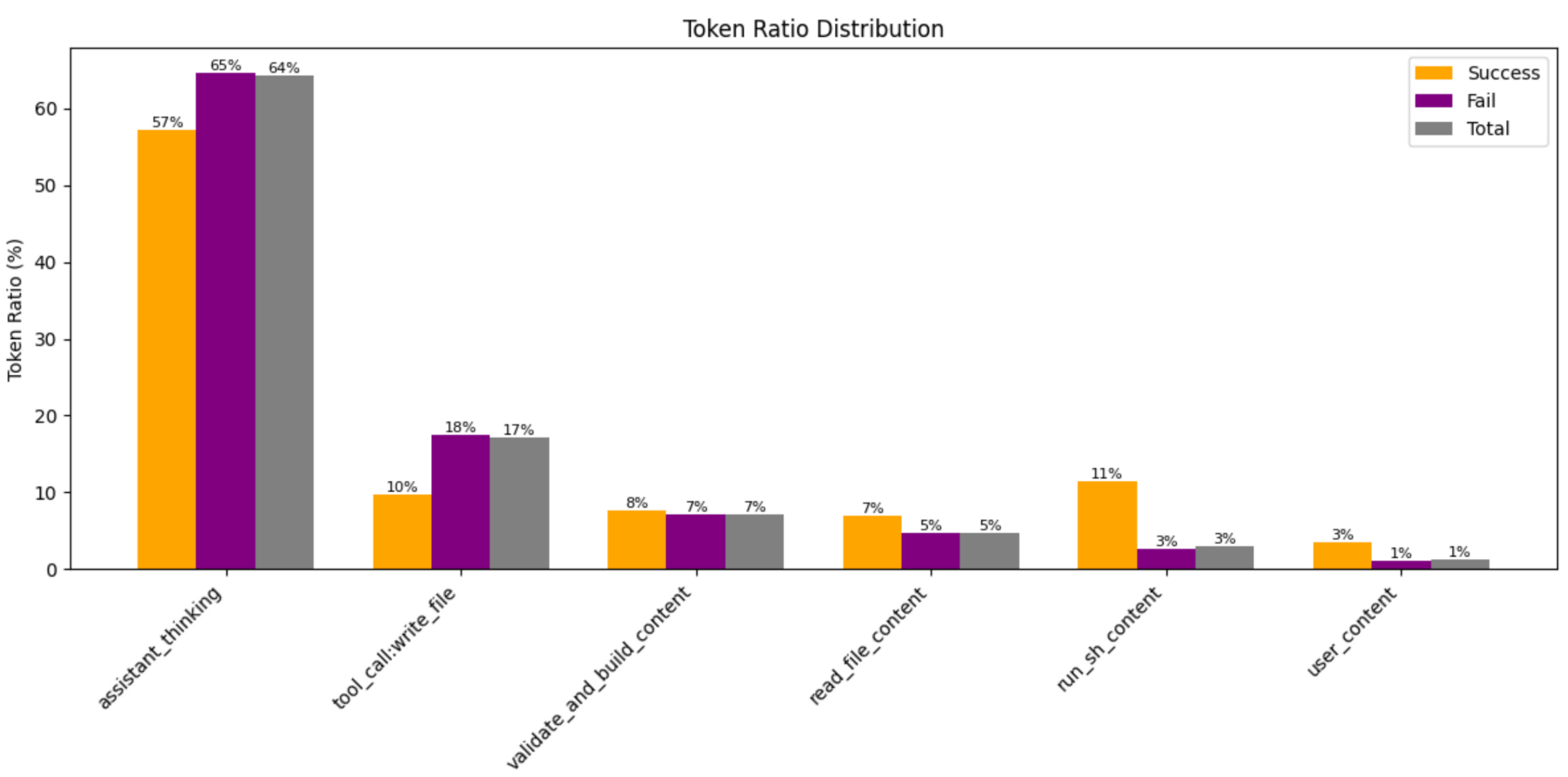}
    \caption{Token distribution across success, failure, and total runs with assistant thinking included.}
    \label{fig:token-ratio-thinking}
\end{figure}

\paragraph{Token Distribution with Thinking Enabled}
\label{app:token_ratio_thinking}
Figure~\ref{fig:token-ratio-thinking} presents the token ratio per trajectory across success, failure, and total runs for the \texttt{Qwen3-32B} base model in the full pipeline. Tokens are grouped into \textit{thinking}, \texttt{tool\_call:write\_file} as the coding is captured within calling the tool rather than its output. For the rest of tools we focus on their output. We also included both user prompt and system prompt (shown as content). We excluded  categories contributing less than 1\%. Over 60\% of all tokens are devoted to internal reasoning (\textit{thinking}), followed by \texttt{write\_file} operations, reflecting the verbosity of full-file generations.  

\begin{figure}[ht]
    \centering
    \includegraphics[width=0.9\linewidth]{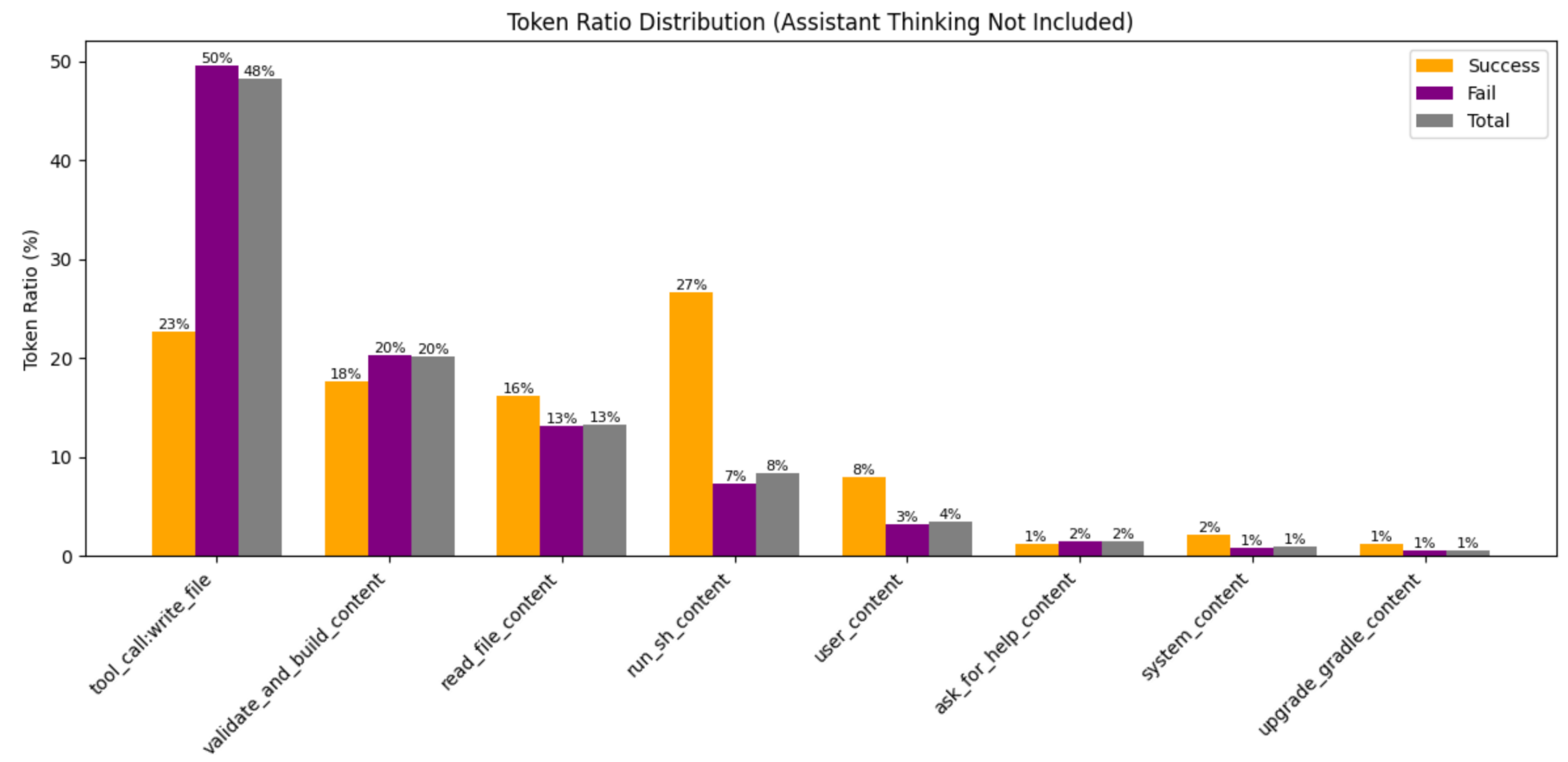}
    \caption{Token distribution excluding assistant thinking, emphasizing tool-level token contributions.}
    \label{fig:token-ratio-nothinking}
\end{figure}

\paragraph{Token Distribution without Thinking}
\label{app:token_ratio_nothinking}
Figure~\ref{fig:token-ratio-nothinking} isolates tool-related contributions by excluding assistant thinking tokens. Majority of tokens used for coding. The distribution highlights that \texttt{run\_sh\_response} accounts for 27\% of tokens in successful runs but only 7\% in failures, whereas \texttt{ask\_for\_help\_response} remains under 2\%, suggesting limited utility of this tool type.  

\begin{figure}[ht]
    \centering
    \includegraphics[width=0.35\linewidth]{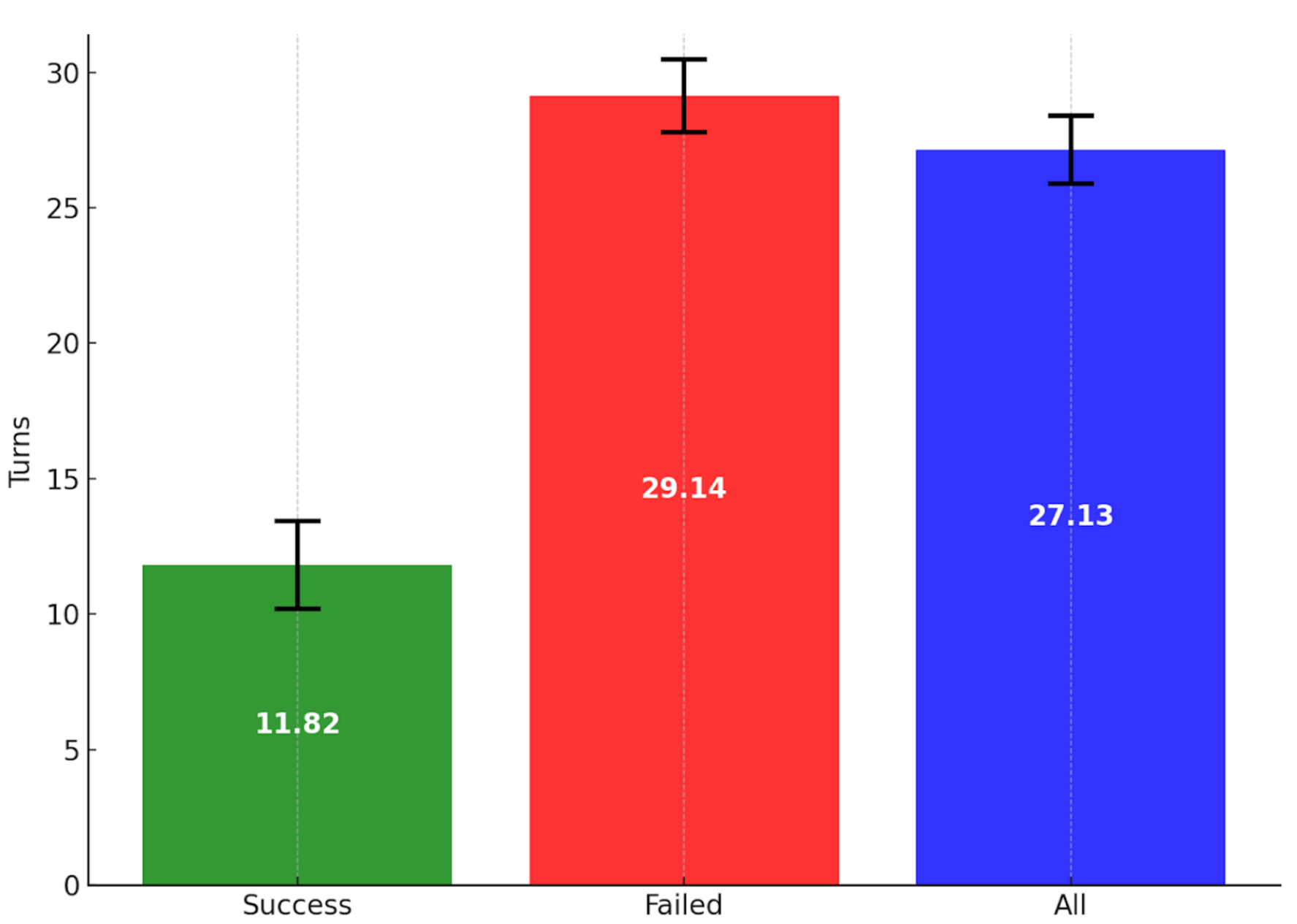}
    \caption{Distribution of trajectory lengths (turns) across successful and failed runs.}
    \label{fig:turns}
\end{figure}

\paragraph{Distribution of Turns per Trajectory}
\label{app:turns}
Figure~\ref{fig:turns} shows the mean $\pm$ 95\% confidence interval of the number of turns per trajectory. Successful repairs typically complete in 12 turns, while failed trajectories average around 30 and frequently reach the 50-step cap, illustrating the efficiency gap between effective and unsuccessful runs.  

\begin{figure}[ht]
    \centering
    \includegraphics[width=0.5\linewidth]{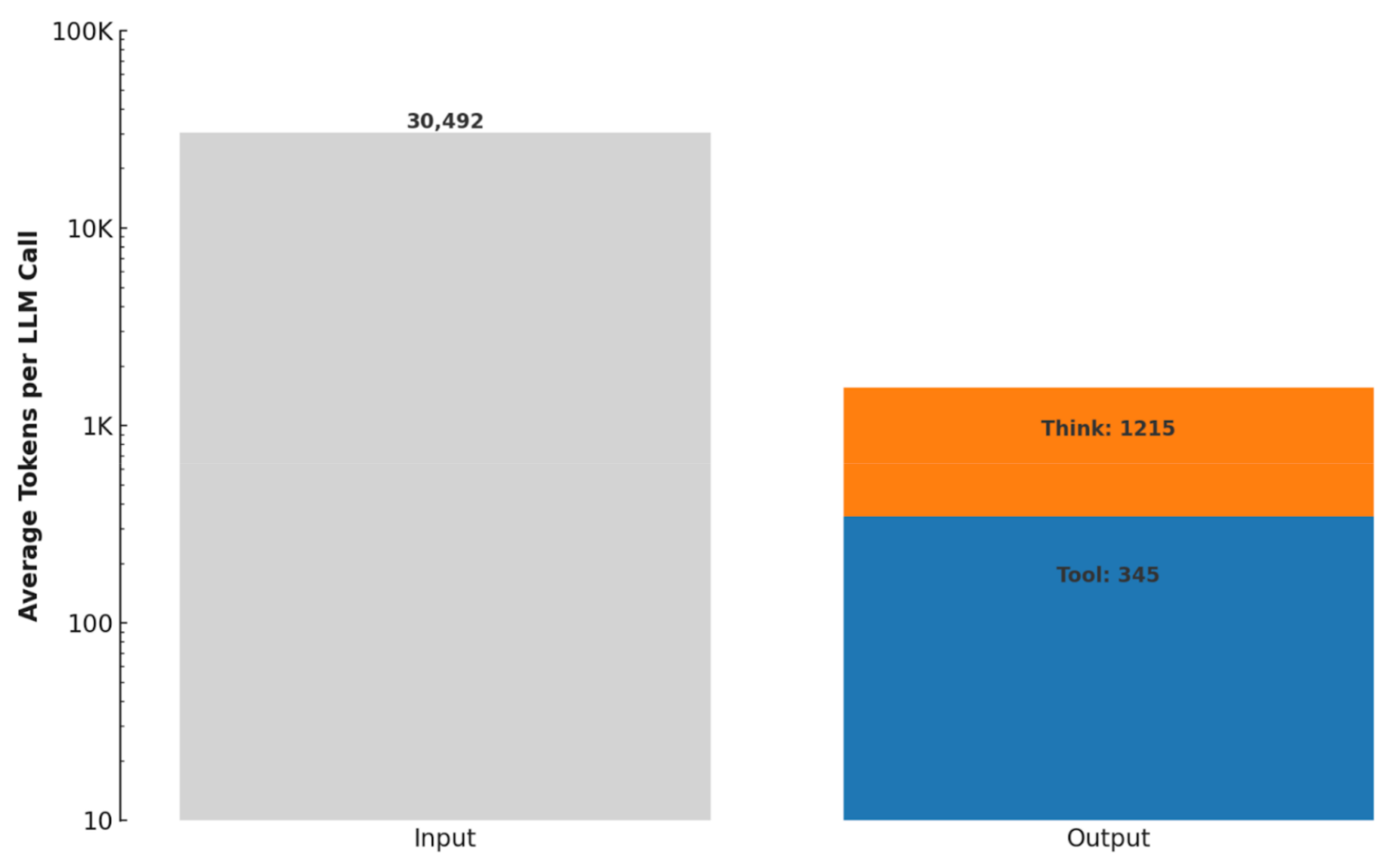}
    \caption{Token utilization per LLM call, showing input/output sizes and distribution across trajectories.}
    \label{fig:token-per-llm}
\end{figure}

\paragraph{Token Utilization per LLM Call}
\label{app:token_per_llm}
Figure~\ref{fig:token-per-llm} reports the input and output token sizes per LLM call on a log scale, capped at 131K tokens due to the \texttt{Qwen3-32B} context limit. Inputs include accumulated prompts, tool responses, and system context, while outputs consist of assistant thinking and tool call tokens. On average, 80\% of output tokens are used for internal reasoning, highlighting the computational overhead of the thinking process.  

\begin{figure}[ht]
    \centering
    \includegraphics[width=0.85\linewidth]{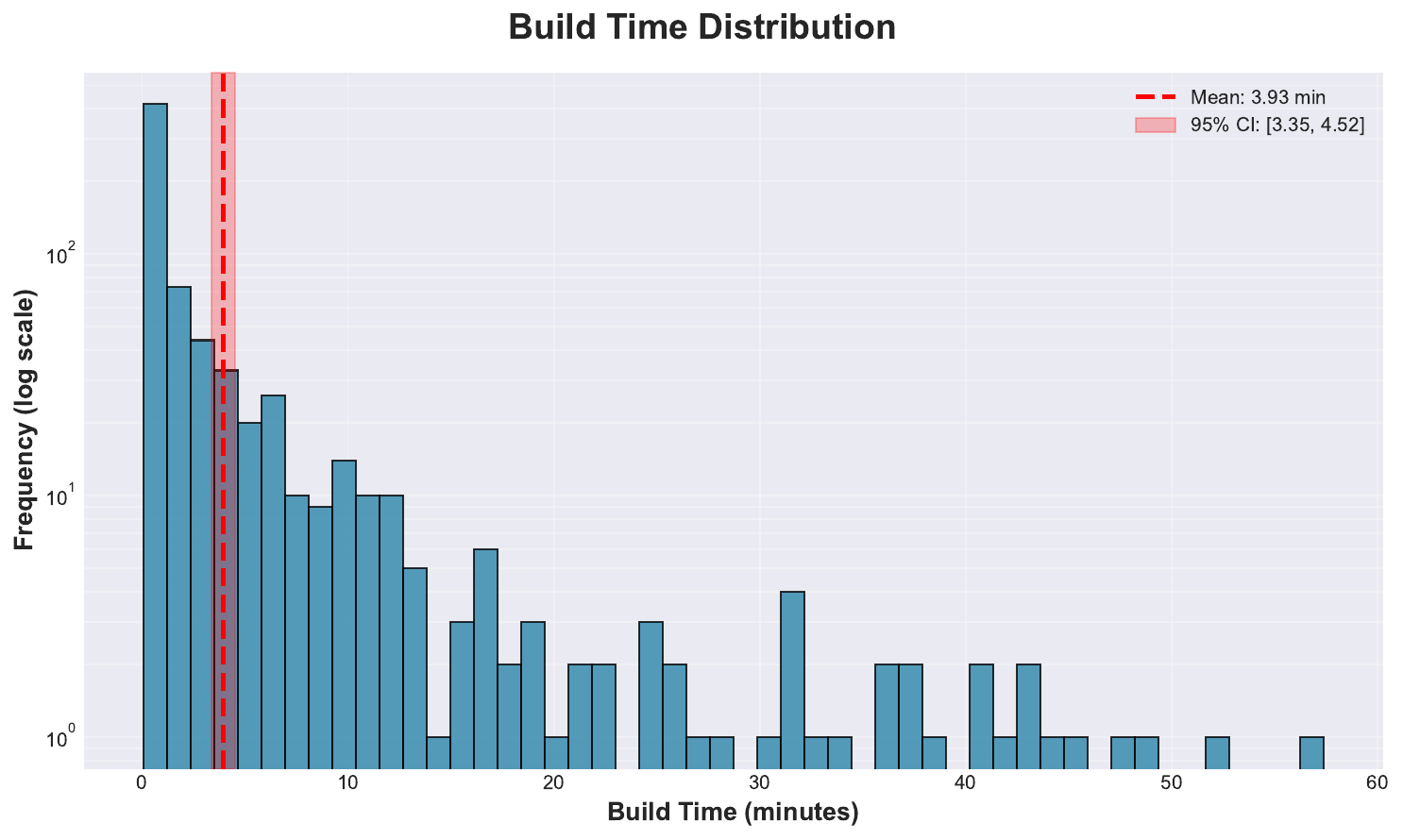}
    \caption{Distribution of build and validation times (log scale) across training problems.}
    \label{fig:build_times_distribution}
\end{figure}

\paragraph{Build and Validation Time Distribution}
\label{app:build_time}
Figure~\ref{fig:build_times_distribution} presents the distribution of build and validation durations across. The build step averaged about $\sim4$~minutes with occasional outliers reaching up to 60~minutes. These long-tail distributions motivated prioritizing repositories with shorter build times during training to maintain high rollout throughput.  

\paragraph{Error Distribution}
\label{app:error_analysis}

To analyze the distribution of build errors in our training dataset, we categorized 2,469 error instances from the simplified pipeline using keyword-based filters. Errors were assigned to categories based on keyword matches, following the rules listed in Table~\ref{tab:error_filters}. Each error is assigned to the first matching rule.

\renewcommand{\arraystretch}{1.25}
\begin{table}[ht]
\centering
\caption{Keyword filters for error categorization. Each error is assigned to the first matching rule.}
\label{tab:error_filters}
\setlength{\tabcolsep}{8pt}
\begin{tabular}{|p{3.5cm}|p{9.0cm}|}
\hline
\textbf{Category} & \textbf{Matching Keywords / Patterns} \\
\hline\hline
\textbf{Dependency-Related} & dependency, dependencies \\
\hline
\textbf{Build Tool} & gradle, maven, build tool, build failed, compilation failed \\
\hline
\textbf{Test} & test, unit test, integration test, test case, test failure \\
\hline
\textbf{Configuration} & configuration, config, schema, avsc, yaml, yml, json, xml \\
\hline
\textbf{Installation} & install, yarn, npm, pip, package manager \\
\hline
\textbf{Version} & version, compatibility, incompatible, mismatch \\
\hline
\textbf{Environment} & path, environment, variable, not found, cannot locate, missing \\
\hline
\textbf{Permission} & permission, access, denied, forbidden \\
\hline
\textbf{Other} & No keyword match above \\
\hline
\end{tabular}
\end{table}

\begin{figure}[ht]
    \centering
    \includegraphics[width=0.75\textwidth]{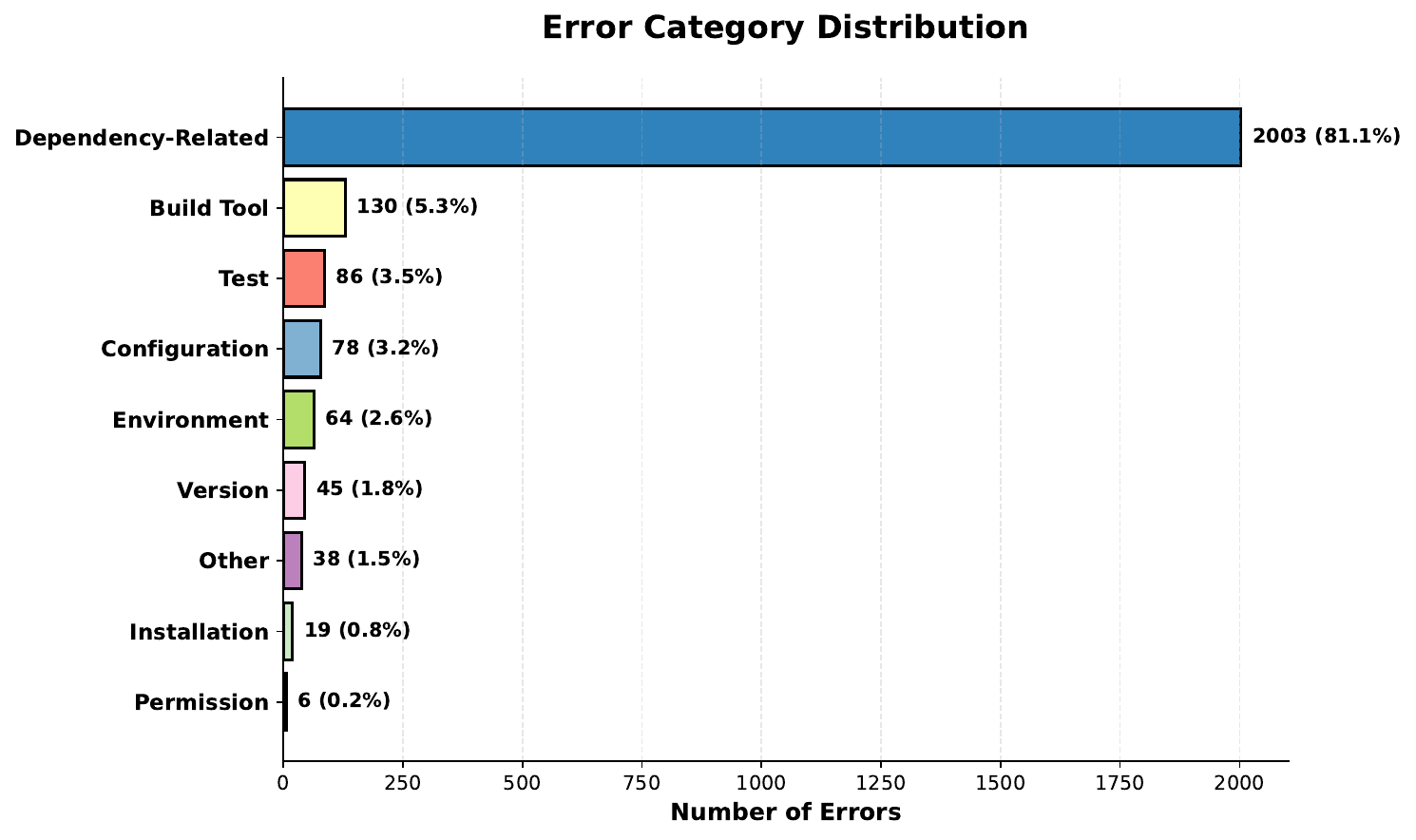}
    \caption{Distribution of error categories in the training dataset of simplified pipeline}
    \label{fig:error_categories}
\end{figure}

Figure~\ref{fig:error_categories} shows the resulting category distribution. We found that \textbf{81.1\%} (2,003 errors) fall under dependency-related issues, including deprecated packages, missing artifacts, duplicate dependencies, and transitive resolution failures. The remaining errors are distributed across Build Tool (5.3\%), Test (3.5\%), Configuration (3.2\%), Environment (2.6\%), Version (1.8\%), Other (1.5\%), Installation (0.8\%), and Permission (0.2\%). These results indicate that dependency management accounts for the majority of build failures in our dataset, making it the most critical target for automated repair.

\end{document}